\documentclass[conf]{new-aiaa}
\usepackage[utf8]{inputenc}

\usepackage{graphicx}
\usepackage{amsmath}
\usepackage[version=4]{mhchem}
\usepackage{siunitx}
\usepackage{longtable,tabularx}
\usepackage[T1]{fontenc}    
\usepackage{hyperref}       
\usepackage{url}            
\usepackage{booktabs}       
\usepackage{amsfonts}       
\usepackage{nicefrac}       
\usepackage{microtype}      

\usepackage{color}
\usepackage{bm}
\usepackage{multirow}
\usepackage{algorithm}
\usepackage{algorithmic}

\title{Benchmarking Long Roll-outs of Auto-regressive Neural Operators for the Compressible Navier-Stokes Equations with Conserved Quantity Correction}

\author{Sean F. Current\footnote{Graduate Researcher, Computer Science and Engineering, current.33@osu.edu} and Chandan Kumar\footnote{Postdoctoral Researcher, Mechanical and Aerospace Engineering, kumar.1144@osu.edu, Member AIAA} and Datta V. Gaitonde \footnote{Glenn Chair Professor, Mechanical and Aerospace Engineering, gaitonde.3@osu.edu, Fellow AIAA} and Srinivasan Parthasarathy\footnote{Professor, Computer Science and Engineering, srini@cse.ohio-state.edu}}
\affil{The Ohio State University, Columbus, Ohio, 43202}

\begin{document}

\maketitle

\begin{abstract}
Deep learning has been proposed as an efficient alternative for the numerical approximation of PDE solutions, offering fast, iterative simulation of PDEs through the approximation of solution operators. However, deep learning solutions have struggle to perform well over long prediction durations due to the accumulation of auto-regressive error, which is compounded by the inability of models to conserve physical quantities. In this work, we present conserved quantity correction, a model-agnostic technique for incorporation physical conservation criteria within deep learning models. Our results demonstrate consistent improvement in the long-term stability of auto-regressive neural operator models, regardless of the model architecture. Furthermore, we analyze the performance of neural operators from the spectral domain, highlighting significant limitations of present architectures. These results highlight the need for future work to consider architectures that place specific emphasis on high frequency components, which are integral to the understanding and modeling of turbulent flows.
\end{abstract}

\section{Introduction}

The usage of machine learning as a methodology for the numerical approximation of solutions to partial differential equations (PDES) has seen significant interest in recent years, owing to the promises of the fast and accurate approximation of solutions in comparison to traditional numerical methods. For complex PDEs that require direct numerical simulation (DNS) to solve, machine learning methods can achieve speed ups of $10^2-10^3$ times \cite{takamoto2022pdebench}, demonstrating significant improvements over their DNS counterparts. However, while these speedups are substantial, the accuracy of machine learning methods can limit their effectiveness at complex problems involving multiple variables or hard conservation criteria.

Recent methods for modeling spatiotemporal PDEs using neural operators \cite{li2020neural, lu2021learning} have achieved significant success at modeling complex PDEs such as the Navier-Stokes equations using short-term auto-regressive predictions \cite{hao2024dpot, hao2023gnot, pathak2022fourcastnet}, but have struggled to maintain performance over longer prediction time frames \cite{hao2024dpot}. This is in large part due to the inability of machine learning methods to conserve fundamental physical quantities as a result of the accumulation of error during auto-regressive prediction \cite{current2024mind}. While error propagation is largely unavoidable for long auto-regressive predictions, the breaking of conservation laws by neural operators accelerates erroneous behavior due to the steady increase or decrease conserved values.

Previous work by Current et al. \cite{current2024mind} has attempted to curtail the breaking of conservation criteria by scaling predicting density terms for the compressible Navier-Stokes equations to maintain mass conservation, however the approach was limited to only testing on the FNO \cite{li2020fourier} and PINO \cite{li2021physics} architectures, and only on CFD datasets over a short time period. In this work, we extend and refine the approach of Current et al. to incorporate conservation terms for momentum, further stabilizing the results of auto-regressive neural operators, and test the ability for operators to predict chaotic flows over significantly longer prediction intervals. In addition, we extend our results to additional neural operator architectures, such as DPOT \cite{hao2024dpot}. Our results demonstrate an advantage for long-term predictions in comparison to baseline models, showing both lower error accumulation over-time and increased durations of high correlation with the ground-truth data. 

\section{Related Work}

Neural Operators \cite{lu2021learning, li2020fourier, li2020neural} have been a recent interest in machine learning for partial differential equations (PDEs), promising efficient and strong approximations of PDE solutions for a wide variety of problems. In contrast to physics-informed neural networks (PINNs) \cite{raissi2019physics, mao2020physics}, which directly learn a function solution to a particular PDE, neural operators instead learn to map between functional solutions to PDEs, often between a initial condition the solution or from one timestep to the next. Neural operators aim to be generalizable to different initial conditions for a PDE, offering a stronger degree of generalizability compared to PINNs, which are trained for specific initial conditions and problem parameters.

Li et al. \cite{li2020fourier} propose the Fourier Neural Operator (FNO), which models data concurrently in the frequency and spatial domains, and has found success model photoacoustic waves \cite{guan2023fourier}, weather modeling \cite{pathak2022fourcastnet}, and a wide variety of smooth PDEs\cite{choubineh2023fourier, li2020fourier, lu2022comprehensive, takamoto2022pdebench}. PINO \cite{li2021physics} incorporates physics-informed losses into the FNO training scheme. Geo-FNO \cite{li2023geometry} apply the Fourier neural operator to no-uniform grids, which is similarly accomplished by NUNO \cite{liu2023nunogeneralframeworklearning}. F-FNO \cite{tran2023factorized} improves the architecture of FNO with additional skip connections and by processing the Fourier modes independently in each spatial dimension. In contrast to the Fourier-based neural operators, DeepONet \cite{lu2021learning, lu2022comprehensive} offers a generalized approach to operator learning, using branch and trunk networks to jointly model the input function and domain space. U-NO \cite{rahman2022u} and CNO \cite{raonić2023convolutionalneuraloperatorsrobust} take inspiration from computer vision architectures and apply U-Net \cite{ronneberger2015unetconvolutionalnetworksbiomedical} architectures to neural operator learning.

More recent approaches have adapted transformer architectures for operator learning, often improving accuracy while reducing the number of model parameters. GNOT \cite{hao2023gnot} offers a general transformer architecture suitable for operator learning which can be applied to a variety of PDE setups. Li et al. propose OFormer \cite{li2023transformerpartialdifferentialequations} and apply transformer operators to learn large-eddy simulations \cite{li2024transformer}. Bryutkin et al. \cite{bryutkin2024hamlet} adapt graph transformers to learn PDEs, and show competitive performance to other models. Further works have constructed pre-trained foundation models for operator learning, such as Poseidon \cite{herde2024poseidon} and DPOT \cite{hao2024dpot}. These approaches have demonstrated stronger stability over long prediction time-frames compared to single-dataset models, but still struggle to model complex turbulence problems over long prediction time frames due to the accumulation of prediction error and chaotic dynamics \cite{hao2024dpot}. These errors can lead to the breaking of fundamental conservation laws, severely limiting the applicability of the models over long prediction regimes. Some prior work has taken steps to reduce conservation errors through hard constraints \cite{current2024mind}, but the methodology was limited only to conservation of mass correction. In this work, we extend upon the work of Current et al. \cite{current2024mind} by incorporating the conservation of momentum into the conservation correction framework, and demonstrate the efficacy of conserved quantity correction on new architectures and complex datasets.

\section{Methods}

\subsection{Problem Setup}

Let us consider the general form of a parametrized time-dependent PDE for variables $\bm{u}(x, t) \in \mathbb{R}^m$, where $(x, t) \in \Omega \times T \subset \mathbb{R}^{d+1}$, where $d$ is the number of spatial dimensions. For a differential operator $F[\bm{u}, \theta](x,t)$ and boundary condition $\mathcal{B}[\bm{u}](x,t)$, the following are satisfied:
\begin{align}
    \frac{\partial \bm{u}}{\partial t} - F[\bm{u}, \phi](x,t) &= 0 \\
    \bm{u}(x, 0) &= \bm{u}^0(x) \\
    \mathcal{B}[\bm{u}](x, t) &= 0,~x \in \partial \Omega,
\end{align}
where $F$ is comprised of spatial derivative terms and $\phi$ are parameters of the PDE. Furthermore, Let $\bm{u}_c(x, t)\in \mathbb{R}^{k\leq m}$ denote the \textit{conserved variables} of $\bm{u}(x, t)$, such that for any time $t\in T$,
\begin{equation}
    \frac{d}{d t}\int_\Omega\bm{u}_c(x, t)~d x = -\int_{\partial \Omega}\bm{J} \cdot \bm{\hat{n}}~dx,
\end{equation}
where $\bm{J}$ is a vector field representing the flux of $\bm{u}(x, t)$ and $\bm{\hat{n}}$ is the unit vector normal to the boundary. In other words, the change in the total amount of the conservative variable $\bm{u}_c$ in the domain $\Omega$ is equal to the total flow of the variable in and out of the domain boundary $\partial\Omega$, assuming no source terms are present within the problem domain. For our purposes, we will focus on closed systems where the net flux is $0$, i.e., 
\begin{equation}
    \frac{d}{d t}\int_\Omega\bm{u}_c(x, t)~d x = 0.
\end{equation}
Understanding the dynamics of conservative variables is integral to many fundamental physical PDEs such as the Navier-Stokes equations, which are the primary focus of this work. By integrating knowledge of conservative variables into state-of-the-art machine learning models for fluid flow prediction, we hope to improve the quality of model predictions as well as stabilize the error of the models over long prediction time frames.

In practice, we work with $\bm{u}$ defined on a discretized domain in both space and time. We represent the solution $\bm{u}$ as a sequence of meshes $\bm{u} = \left\{\bm{u}^0, \bm{u}^1,...,\bm{u}^T\right\}$. For this work, we will assume all meshes are defined on a regular grid $\mathcal{X}$ such that $\bm{u}^t$ can be represented as a real-valued $d$-dimensional tensor with $d$ indices. While irregular grids are commonly used to represent PDE solutions, many existing works have proposed techniques to transform irregular geometries into regular grids for the purpose of machine learning applications \cite{hao2024dpot, li2023geometry}, thus we primarily focus on regular geometries for this work.

\subsection{Auto-regressive Modeling}

We aim to learn a conservative, auto-regressive operator $\mathcal{H}_\theta$ parameterized by learnable weights $\theta$ that predicts a future timestep $\bm{u}^{t+1}$ conditioned on a subset of $h$ previous timesteps $\bm{u}^{t},...,\bm{u}^{t-h+1}$:
\begin{equation}
    \bm{u}^{t+1} = \mathcal{H}_\theta\left(\bm{u}^{t},...,\bm{u}^{t-h+1}\right).
\end{equation}
Prior works have utilized an extensive temporal history to predict the next timestep: FNO \cite{li2020fourier} and DPOT \cite{hao2024dpot} both use 10 timesteps as a history for the time-dependent Navier-Stokes datasets, which is nearly half of the timesteps available in the commonly used PDEBench dataset \cite{takamoto2022pdebench}. In contrast, Tran et al. \cite{tran2023factorized} find that using a single timestep as history is sufficient, directly mimicking the approach of direct numerical solvers. We follow the approach of Tran el al. and use only a single prior timestep as input to our model, constructing the basic recursion scheme
\begin{equation}
    \bm{u}^{t+1} = \mathcal{H}_\theta\left(\bm{u}^{t}\right).
\end{equation}

Many techniques have been proposed to train auto-regressive models, ranging from auto-regressive rollout, teacher forcing, the push-forward trick \cite{brandstetter2022message}, and denoising \cite{hao2024dpot}. Unlike auto-regressive rollouts and the push-forward trick, teacher forcing and denoising do not require recursive passes through the model during training, greatly improving their efficiency. While efficient, these methods do not enforce recurrent relations in the model, which may impact performance during inference due to the accumulation of small prediction errors. Furthermore, the denoising strategy \cite{hao2024dpot} breaks conservation constraints through the addition of noise to the input, limiting the ability of the model to learn properties of conservation. We opt to train our models using the auto-regressive rollout strategy over a limited number of timesteps $\tau$: by constraining the model to only predicting a short duration, we limit the explosion of error terms, stabilizing the loss gradient while still training effectively in an auto-regressive manner. We utilize the $L_2$ relative error as a loss function, calculated individually for each timestep:
\begin{equation}
    \label{ch5.eq:loss}
    \mathcal{L} = \sum_{i=0}^\tau\frac{\left|\left|\mathcal{H}_{\theta}^i(\bm{u}^t) - \bm{u}^{t + i}\right|\right|_2}{\left|\left|\bm{u}^{t + i}\right|\right|_2},
\end{equation}
where the $L_2$-norm $||\cdot||_2$ of the tensor $\bm{u}^t$ is defined as $\sqrt{\sum_{i_1,...i_d}{\bm{u}_c}^{t}(i_1,...,i_d)^2}$. Using the $L_2$ relative error helps balance the loss across different variable scales, while still allowing the model to maintain conserved quantities, which may not be preserved with commonly used techniques such as normal or min-max scaling.

\subsection{Model Architecture}
\label{ch5.sec:architecture}

\begin{figure}[h!]
    \centering
    \includegraphics[alt={A schematic of the DPOT model with conserved quantity correction.},width=0.9\linewidth]{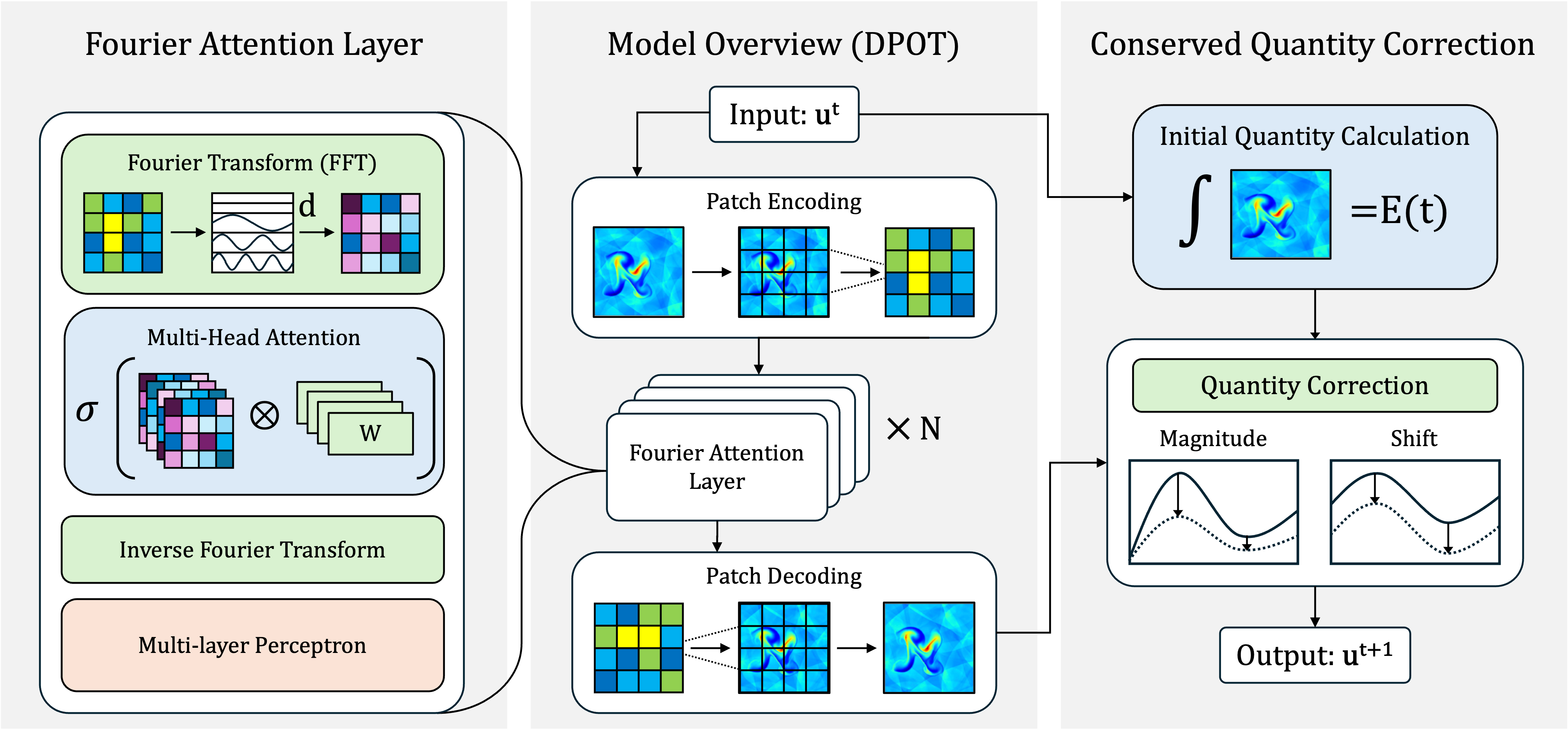}
    \caption{An example of the modeling framework for the DPOT architecture. Left: the Fourier Attention Layer, comprised of a Fourier transform, multi-head block attention, inverse Fourier transform, and an MLP. Middle: the general model architecture, including patch encoding, Fourier Attention Layers, and patch decoding. Right: the Conserved Quantity Correction Scheme, which calculates the initial quantity and corrects the model output via magnitude or shift scaling. A similar framework is used for FNO.}
    \label{ch5.fig:schema}
\end{figure}

\begin{figure}[h!]
    \centering
    \includegraphics[alt={A schematic of the FNO model.},width=0.9\linewidth]{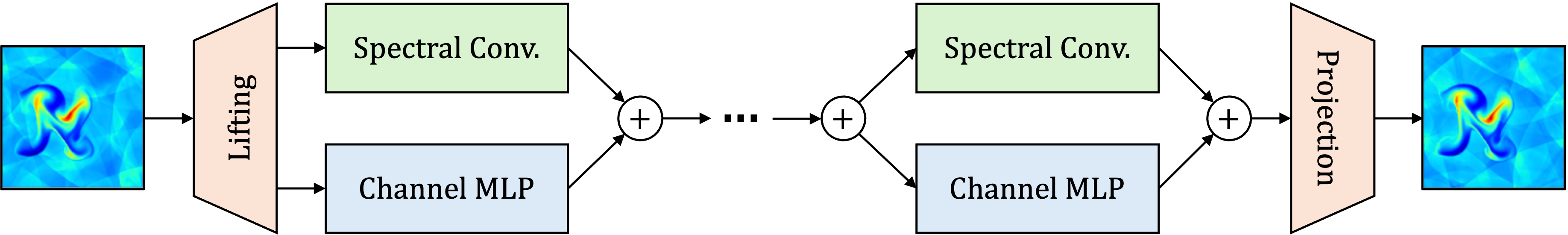}
    \caption{A schematic of the FNO architecture, including lifting, spectral convolutions, channel-wise MLPs, and projection operations. This architecture can be used in place of the DPOT architecture in Figure \ref{ch5.fig:schema}.}
    \label{ch5.fig:fno_schema}
\end{figure}

To improve the generalizability of our methodology, we construct $\mathcal{H}_\theta$ as a conservative transformation $\mathcal{T}$ of a learnable model $\mathcal{G}_\theta$, such that 
\begin{equation}
    \mathcal{H}_\theta(\bm{u}^{t}) = \mathcal{T}(\mathcal{G}_{\theta})(\bm{u}^{t}),
\end{equation}
where $\mathcal{T}(\mathcal{G}_\theta)$ is the operation of $\mathcal{T}$ on $\mathcal{G}_\theta$. This allows the architecture of $\mathcal{G}_\theta$ to be separated from the conservation transformation, allowing a strong degree of flexibility in the construction of $\mathcal{G}_\theta$. In this work, we primarily focus on two architectures for $\mathcal{G}_\theta$: FNO \cite{li2020fourier} and DPOT \cite{hao2024dpot}, without loss of generality. The Fourier Neural Operator (FNO) primarily learns via spectral convolution, which applies the Fast Fourier Transform (FFT) to data and multiplies the resulting Fourier modes by learnable weight matrices before converting data back to the spatial domain. A visualization of the FNO architecture is provided in Figure \ref{ch5.fig:fno_schema}. The DPOT architecture similarly processes data in the Fourier space, but does so in a attention-based manner by applying a set of learned weight matrices to groups of embedded channels. The DPOT architecture also utilizes patch embedding techniques to better incorporate local dynamics. A visualization of the DPOT architecture is provided in Figure \ref{ch5.fig:schema}.

Our implementation of the DPOT architecture differs from the original implementation by a reduction of the temporal history. In the original work, 10 prior timesteps are used as input to the model. We find such an extensive history to be unnecessary for training DPOT on singular datasets while still maintaining strong predictive capabilities. Our models use only the previous two timesteps as input to the model: we find that the inclusion of more than one timestep promotes long-term stability and reduces autoregressive errors, but the inclusion of more than two timesteps does not have significant impact on the model performance, and could limit real-world application of models if the temporal history is too extensive. Furthermore, by reducing the need for a large set of prior timesteps as input, we can use additional timesteps from the dataset to train our models, which would normally be reserved as input features for models with a long history. For most datasets, the initial timesteps are often the most structurally dynamic, as data slowly becomes more stationary over time. By allowing our models to train on the earlier dynamic timesteps we may also improve model performance on the whole.

In addition, we incorporate the internal normalization scheme used by Huang et al. \cite{huang2017arbitrary} for the DPOT model, which has previously been applied by Li et al. \cite{li2023geometry} for operator learning. This technique balances the impact of different variable scales by normalizing data while also allowing the model to learn changes in the scale of variables. Additionally, because the normalized scaling of variables may make it difficult for models to learn hard variable limits (such as 0 for density or pressure), we clip any predicted values with a minimum of 0 to $1\times 10^{-8}$. This prevents the prediction of negative values which may accentuate the error of models during auto-regressive prediction.

\subsection{Conserved Quantity Correction}
\label{ch5.sec:cqc}

To construct the conservative transformation operator $\mathcal{T}$, we must recognize two subsets of possible conserved variables $\bm{u}_c$, based on the range of possible values $\bm{u}_c(x, t)$ can take, which changes how the conservation law can be guaranteed.

\paragraph{Magnitude Correction.} In the first case, the variable $\bm{u}_c(x, t)$ is a scalar quantity such that $\forall~x\in \mathcal X$ and $t\in T$, $\bm{u}_c(x, t) \geq 0$, such as in the case of density or energy density terms. This constraint ensures that the integral
\begin{equation}
    \int_\Omega \bm{u}_c(x, t)~dx \geq 0
\end{equation}
for all $t$. Thus, we can scale the output of $\mathcal{G}_\theta$ using information from $\bm{u}^{t-1}$ to ensure the net change in the conserved quantity is $0$:
\begin{align}
    \mathcal{H}_\theta(\bm{u}_c^{t}) &= \mathcal{T}(\mathcal{G}_\theta)(\bm{u}_c^{t}) \\
    &= \mathcal{G}_\theta(\bm{u}_c^{t})\left(\frac{\int_\Omega\bm{u}_c^{t}~dx}{\int_\Omega\mathcal{G}_\theta(\bm{u}_c^{t})~dx}\right).
\end{align}
Because $\bm{u}^{t}$ is represented discretely as a tensor, the integral terms can be approximated by summing over the tensor entries using the $l_1$-norm, where $||\bm{u}_c^{t}||_1 = \sum_{i_1,...i_d}\left|{\bm{u}_c}^{t}(i_1,...,i_d)\right|$:
\begin{equation}
    \mathcal{H}_\theta(\bm{u}_c^{t}) = \mathcal{G}_\theta(\bm{u}_c^{t})\frac{||\bm{u}_c^{t}||_1}{||\mathcal{G}_\theta(\bm{u}_c^{t})||_1}.
\end{equation}
 This scaling ensures that the output quantity $\bm{u}^{t+1}\approx \mathcal{H}_\theta(\bm{u}_c^{t})$ is conserved based on the input quantity $\bm{u}_c^{t}$, and guarantees that $\mathcal{H}_\theta(\bm{u}_c^{t}) \geq 0$ if the output of the learned model $\mathcal{G}_\theta(\bm{u}_c^{t}) \geq 0$. This process is akin to the scaling procedure used in Current et al. \cite{current2024mind}.

\paragraph{Shift Correction.} In the second case, the variable $\bm{u}_c(x, t)$ is a one-dimensional vector quantity that may take on negative values, such as the case of momentum terms. In this case, we cannot ensure that the conserved quantity
\begin{equation}
    \int_\Omega \bm{u}_c(x, t)~dx \neq 0,
\end{equation}
and in many experiments, the integral may be explicitly expected to equal $0$. Furthermore, for two tensors $\bm{u}_c^t$ and $\bm{u}_c^{t^\prime}$, it may be the case that 
\begin{equation}
    \int_\Omega \bm{u}_c^t~dx = \int_\Omega \bm{u}_c^{t^\prime}~dx
\end{equation}
but
\begin{equation}
    ||\bm{u}_c^{t}||_1 \neq ||\bm{u}_c^{t^\prime}||_1
\end{equation}
due to the presence of negative directional components in $\bm{u}_c$. Thus, we cannot utilize the scaling procedure used for scalar quantities. Instead, we use an alternative scaling procedure to construct $\mathcal{H}_\theta$:
\begin{align}
    \mathcal{H}_\theta(\bm{u}_c^{t}) &= \mathcal{T}(\mathcal{G}_\theta)(\bm{u}_c^{t}) \\
    &= \mathcal{G}_\theta(\bm{u}_c^{t}) + \left(\frac{\int_\Omega\bm{u}_c^{t}~dx - \int_\Omega\mathcal{G}_\theta(\bm{u}_c^{t})~dx}{A}\right),
\end{align}
where $A$ is the area of the domain $\Omega$. As before, we can use the sum of the tensor components to approximate the integral terms. Like the prior scaling method, this transformation ensures that the output quantity $\mathcal{H}_\theta(\bm{u}_c^{t})$ is conserved based on the input quantity $\bm{u}_c^{t}$. Unlike the scalar transformation, however, the vector scaling allows for negative output values, maintaining the directional utility of $\bm{u}_c$.

\section{Experiments}

We apply our methodology to the 2-D Compressible Navier-Stokes Equations as supplied by PDEBench \cite{takamoto2022pdebench}. The 2-D compressible Navier-Stokes equations are defined by the density $\rho(x, t)$, pressure $p(x, t)$, and velocity $\bm{u}(x, t)$ fields following the PDEs
\begin{align}
    \partial_t\rho + \nabla \cdot(\rho\bm{u}) &= 0, \\
    \rho(\partial_t\bm{u} + \bm{u}\cdot\nabla\bm{u}) &= -\nabla p + \eta\Delta\bm{u} + (\zeta + \eta/3)\nabla(\nabla\cdot\bm{u}), \\
    \partial_t\left(\frac{3}{2}p+\frac{\rho u^2}{2}\right) &= -\nabla\cdot\left(\left(\epsilon + p + \frac{\rho u^2}{2}\right)\bm{u} - \bm{u} \cdot \sigma^\prime\right),
\end{align}
where $\eta$ is the shear viscosity, $\zeta$ is the bulk viscosity, $\sigma$ is the viscous stress tensor, and $\epsilon$ is the internal energy of the system. We focus on the inviscid datasets with viscosity $\eta=\zeta=10^{-8}$ and mach number $M \in \{0.1, 1.0\}$. These datasets were chosen for their highly dynamic nature: while PDEBench provides datasets with higher viscosity ($\eta,\zeta\in\{10^{-1},10^{-2}\}$), these datasets tend to be rather static over longer durations, resulting in the learning of trivial near-identity functions for some channels. By selecting low viscosity, high motion datasets, we can better understand the long-term capabilities of the varying model architectures. 

Additionally, while prior work has focused primarily on modeling the density, pressure, and velocity terms directly, we instead opt to model the solution using the conserved quantities density $\rho(x,t)$, momentum $\rho\bm{u}(x, t)$, and the energy $E(x, t) = \frac{3}{2}p +\frac{\rho u^2}{2}$ for models using conserved quantity correction. This form maximizes the number of conserved variables we can apply conserved quantity correction to, while also resembling the terms most commonly modeled within numerical solvers. Notably, the total density and momentum are constant throughout time for the specified PDEs, leaving only the energy term un-conserved, which may fluctuate due to the dissipation of energy into heat due to viscosity or the conversion of internal energy into kinetic. Thus, conserved quantity correction is applied to the mass and momentum terms while the energy is left uncorrected. To ensure fair comparisons to benchmark methods, all predictions are converted back to primitive values (density, velocity, and pressure) before calculating error metrics.

\subsection{Reproducibility}

We train our models using an Adam optimizer with a one-cycle learning rate scheduler over 100 epochs, with the first 20 epochs reserved for the warm-up period. We use an initial learning rate of $1\times10^{-3}$ and default momentum parameters $\beta_1=0.9,\beta_2=0.999$. All experiments use a batch size of 16. We report results for baseline DPOT and FNO models, DPOT and FNO models with mass scaling according to Current et al. \cite{current2024mind}, denoted with a $_m$, and models with our conserved quantity correction, denoted with a $_c$. We set the patch size of the DPOT architecture to 4 and the number of modes for the DPOT and FNO models to 32, which we find works well for the high-frequency features observed in the CFD data. All datasets are downsampled to a $128 \times 128$ grid following Hao et al. \cite{hao2024dpot}. Models are trained auto-regressively over 5 timesteps using the loss in Equation \ref{ch5.eq:loss}. We also include results for two additional DPOT baselines, DPOT-Ti and DPOT-S, which follow the parameter settings and training methodology from their original implementation by Hao et al. \cite{hao2024dpot}. All experiments are run on a single Nvidia RTX A6000 GPU.

\subsection{Results}

\begin{table}[h!]
    \footnotesize
    \centering
    \caption{L2 relative error for all experiments. The best model for each timestep is \textbf{bolded}.}
    \label{ch5.tab:results}
    \begin{tabular}{c c c c c c c c c c}
        \toprule
        \multirow{2}{*}{Model} & \multirow{2}{*}{Params} & \multicolumn{4}{c}{$[\eta, \zeta]=10^{-8}, M=0.1$} & \multicolumn{4}{c}{$[\eta, \zeta]=10^{-8}, M=1.0$} \\
        & & Avg. Err & T=1 & T=5 & T=10 & Avg. Err & T=1 & T=5 & T=10 \\
        \hline
        baselines & & & & & & & & & \\
        FNO & 36M & 0.119 & 0.075 & 0.113 & 0.148 
                  & 0.252 & 0.132 & 0.230 & 0.345 \\
        FNO$_m$& 36M & 0.109 & 0.069 & 0.104 & 0.137 
                     & 0.248 & 0.129 & 0.225 & 0.342 \\
        DPOT & 4.8M & 0.069 & 0.049 & 0.065 & 0.084 
                    & 0.151 & 0.086 & 0.132 & 0.210 \\
        DPOT$_m$ & 4.8M & 0.067 & 0.047 & 0.064 & 0.083 
                        & 0.150 & 0.086 & 0.131 & 0.208 \\
        DPOT-Ti & 7.5M & 0.093 & 0.054 & 0.088 & 0.118 
                       & 0.156 & 0.084 & 0.136 & 0.220 \\
        DPOT-S & 31M & 0.088 & 0.052 & 0.084 & 0.113 
                     & \textbf{0.141} & \textbf{0.072} & \textbf{0.123} & \textbf{0.129} \\
        \hline
        FNO$_c$ & 36M & 0.098 & 0.063 & 0.094 & 0.122 
                      & 0.180 & 0.107 & 0.162 & 0.243 \\
        DPOT$_c$ & 4.8M & \textbf{0.064} & \textbf{0.045} & \textbf{0.061} & \textbf{0.078} 
                        & 0.151 & 0.089 & 0.132 & 0.209 \\
        \bottomrule
    \end{tabular}
\end{table}

The results for the PDEBench datasets over 10 timesteps are presented in Table \ref{ch5.tab:results}, which displays the relative $L_2$ error for timesteps +1, +3, +5, and +10, predicted autoregressively. To maintain fair error calculations, all rollouts are seeded at time $t=10$ to account for the temporal history requirements of DPOT-Ti and DPOT-S. As seen in the table, the models with conserved quantity correct achieve consistently lower error than their alternative models on the Mach 0.1 dataset, which becomes increasingly prominent as the timestep grows larger. the DPOT$_c$ model is able to achieve a relative error of only $7.8\%$ after 10 auto-regressive predictions, compared to the $8.3\%$ error of the baseline DPOT and DPOT$_m$ models. In contrast, the DPOT models with conserved quantity correction are not significantly better at predicting the Mach 1.0 dataset, performing comparably to their un-conservative counterparts. Notably, the FNO$_c$ model exhibits significant performance improvements on the Mach 1.0 dataset compared to the baselines, demonstrating drastic reductions in error compared to FNO and FNO$_m$. This improvement is not observed in the DPOT models. All models perform significantly worse at the Mach 1.0 dataset, which is a far more difficult problem compared to the slower flow of Mach 0.1. The Mach 1.0 dataset exhibits significantly more high frequency features which the smaller models tested in this work are less capable of modeling with fidelity. The larger DPOT-S model demonstrates stronger performance on the Mach 1.0 dataset, suggesting that further improvements could be obtained using larger models.

\subsection{Channel-Specific Performance}

Because the compressible Navier-Stokes equations require accurate modeling of multiple different variables, we aim to not only characterize the overall error of our models, but also consider variable-specific errors and how they may contribute to the overall error of the models. The average relative error for each variable over the entire prediction time-frame is reported in Table \ref{ch5.tab:indv_results}. For all models, the relative error of the velocity term is significantly larger than the error terms for density or pressure, often on the order of 4-5 times larger compared to density and 8-10 times larger compared to pressure. However, while the error of the velocity terms is larger, the correlation between the model output and the velocity terms remains high throughout the prediction, often maintaining a correlation of $0.99$ even after 10 predictions. In contrast, the correlation for the density channel decays more aggressively despite having a lower relative error, shifting on average from $0.99$ for $t=1$ to $0.95$ for $t=10$ for the DPOT$_c$ model. This discrepancy may be attributed to the inability of neural operators to attend to finer scale features: the scale of the velocity terms for the data is 1-2 magnitudes lower than the density or pressure terms. While the variable scale is accounted for by the relative error metric, the neural operators themselves may be less effective at operating across variables at different scales, despite the inclusion of feature normalization layers as discussed in Section \ref{ch5.sec:architecture}.

\begin{table}[h!]
    \small
    \centering
    \caption{L2 relative error per channel for all experiments. The best model for each channel is \textbf{bolded}.}
    \label{ch5.tab:indv_results}
    \begin{tabular}{c c c c c c c c}
        \toprule
        \multirow{2}{*}{Model} & \multirow{2}{*}{Parameters} & \multicolumn{3}{c}{$[\eta, \zeta]=10^{-8}, M=0.1$} & \multicolumn{3}{c}{$[\eta, \zeta]=10^{-8}, M=1.0$} \\
        & & $\rho$ & $p$ & $\bm{u}$ & $\rho$ & $p$ & $\bm{u}$ \\
        \hline
        baselines  & & & & & & \\
        FNO & 36M & 0.046 & 0.202 & 0.024 & 0.097 & 0.405 & 0.103 \\
        FNO$_m$& 36M & 0.043 & 0.187 & 0.021 & 0.093 & 0.400 & 0.101 \\
        DPOT & 4.8M & 0.029 & 0.116 & 0.013 & 0.067 & 0.236 & 0.065 \\
        DPOT$_m$ & 4.8M & 0.028 & 0.114 & 0.013 & 0.067 & 0.234 & 0.064 \\
        DPOT-Ti & 7.5M & 0.040 & 0.157 & 0.017 & 0.070 & 0.244 & 0.066 \\
        DPOT-S & 31M & 0.036 & 0.151 & 0.016 & \textbf{0.065} & \textbf{0.219} & \textbf{0.060} \\
        \hline
        FNO$_c$ & 36M & 0.041 & 0.167 & 0.018 & 0.078 & 0.284 & 0.075 \\
        DPOT$_c$ & 4.8M & \textbf{0.027} & \textbf{0.108} & \textbf{0.012} & 0.067 & 0.236 & 0.064 \\
        \bottomrule
    \end{tabular}
\end{table}

\subsection{Long-term Rollouts}

To test the efficacy of our models for long-term predictions, we generate an additional 50 time steps for 100 samples from the 2D Compressible flow dataset with $\eta,\zeta=10^{-8}$ and $M=0.1$. Beyond 50 timesteps, the flows begin to exhibit highly chaotic behavior, and all methods begin to exhibit high rates of error as high frequency components begin to become more prominent, thus we restrict our analysis to the range $[0, 50]$. Note that we do not retrain models on the extended duration: models are only evaluated on the long-term dataset. This lets us evaluate model performance when encountering unseen dynamics over long prediction durations.

\begin{figure}[h!]
    \centering
    \includegraphics[alt={A plot showing different error rates as a function of time.},width=0.5\linewidth]{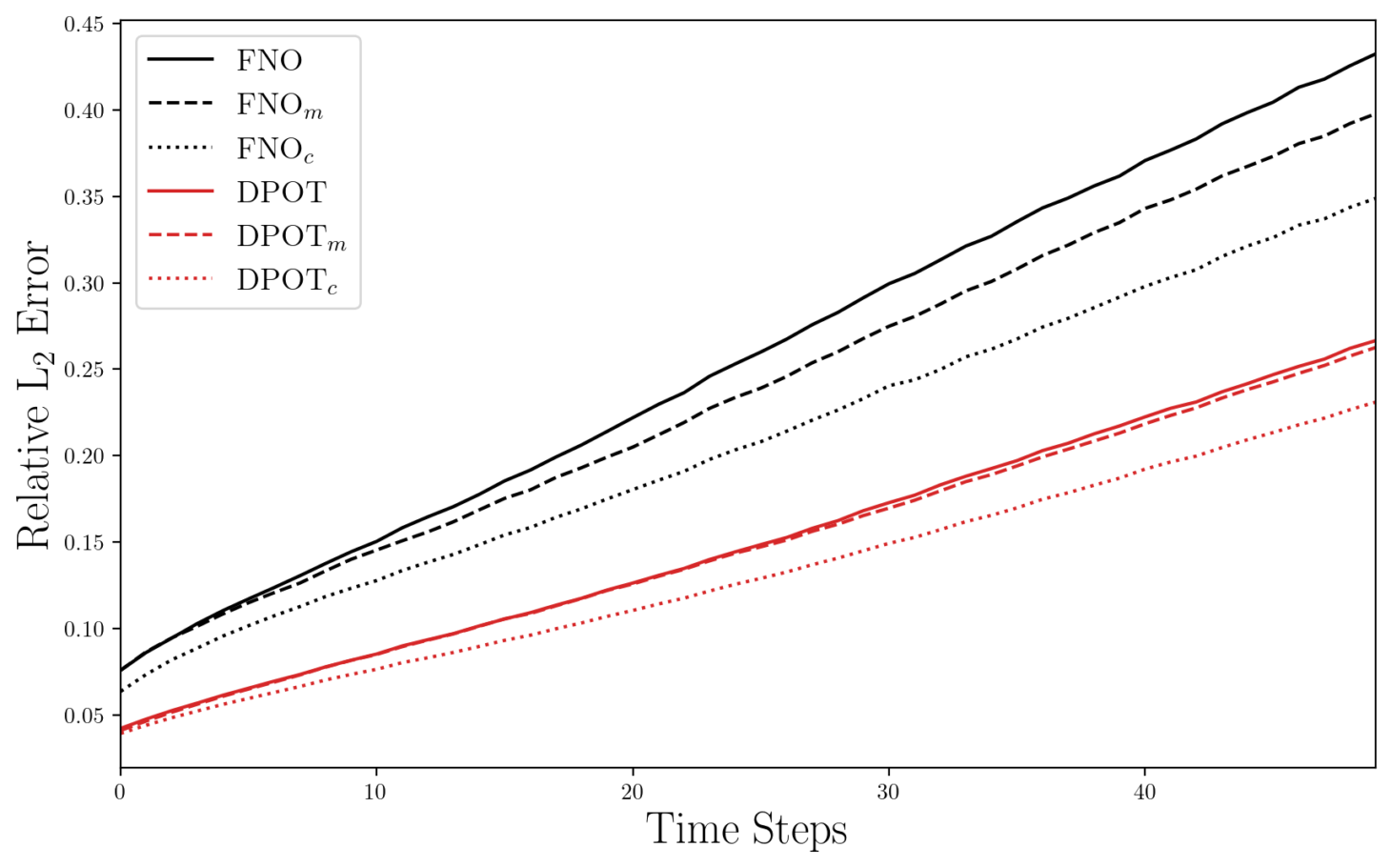}
    \caption{$L_2$ relative error for models over a long prediction interval. The models with mass and momentum quantity correction (denoted with a $c$ subscript) consistently achieve lower error than their baseline and mass conserving (denoted with a $m$ subscript) counterparts.}
    \label{ch5.fig:long_loss}
\end{figure}

The $L_2$ error of the long rollout results is displayed in figure \ref{ch5.fig:long_loss}. For all modeling schemes, the model with conserved quantity correction maintains the lowest error throughout the entire prediction regime, with the relative error increasing by only $0.4\%$ per timestep for the DPOT$_c$ model. In addition, the output of DPOT$_c$ exhibits a high correlation with the ground-truth data for a longer period of time than other models, achieving an average duration of $28$ timesteps before the correlation coefficient falls beneath $90\%$. In contrast, the base DPOT model achieves a high correlation for only $25$ timesteps, and the FNO models achieve high correlation for only $13$. This demonstrates that the models with conserved quantity correction are better at learning and progressing flow structures than their unconstrained counterparts, despite quantity correction being applied uniformly throughout the problem domain.

\begin{figure}[h!]
    \centering
    \includegraphics[alt={An array of images showing the final prediction for different models.},width=0.8\linewidth]{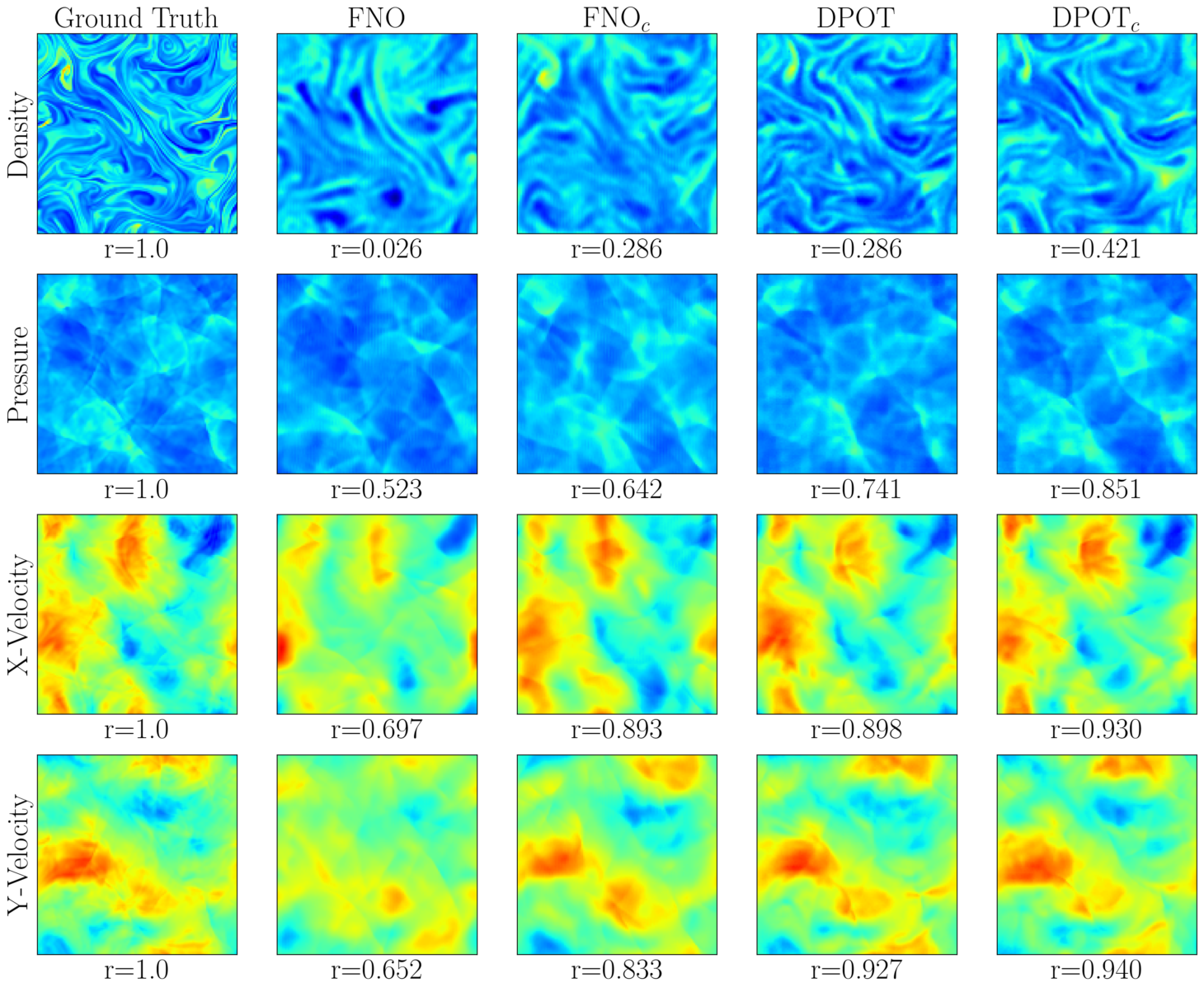}
    \caption{Final timestep predictions for the FNO, FNO$_c$, DPOT, and DPOT$_c$ models for a long rollout test sample. $r$ indicates the correlation between the predicted sample and the ground truth.}
    \label{ch5.fig:last_timestep}
\end{figure}

\subsection{Spectral Analysis}

Finally, we analyze the differential performance of our models from the perspective of the frequency domain. We primarily focus this analysis on two features of the flow which exhibit notable characteristics from the frequency perspective: the fluid density $\rho$, which demonstrates significant shift in frequency spectra as a function of time, and the turbulent kinetic energy (TKE), calculated as TKE$=\frac{1}{2}\rho\bm{u}^2$, which exhibits a known decay in the frequency regime due to turbulent energy cascade \cite{richardson1922weather, kolmogorov1941local}. Energy cascade describes the phenomena in which energy is transferred from larger flow structures to smaller ones, which eventually dissipate kinetic energy into heat via viscous forces. The transfer of energy from low frequency to high frequency regimes should occur at a near constant exponential rate, dependent on the kinematic viscosity of the flow.

\begin{figure}[h!]
    \centering
    \includegraphics[alt={A spectral density plot showing the distribution of kinetic energy over frequencies.},width=0.8\linewidth]{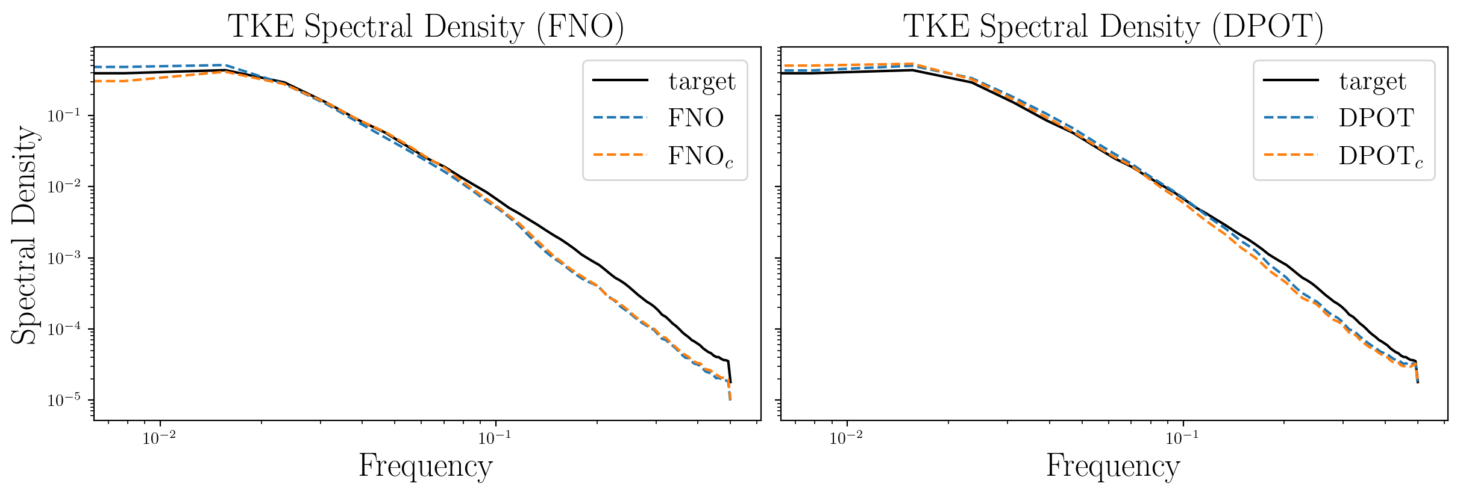}
    \caption{TKE spectra for an example from the Mach 0.1 dataset. FNO-based models are presented in the left plot, while DPOT-based models are presented on the right. While both models exhibit lower density than the ground-truth for high frequency modes, the DPOT models show a lesser difference than the FNO models.}
    \label{ch5.fig:tke}
\end{figure}

\begin{figure}[h!]
    \centering
    \includegraphics[alt={Frequency spectra for the density channel over time.},width=0.9\linewidth]{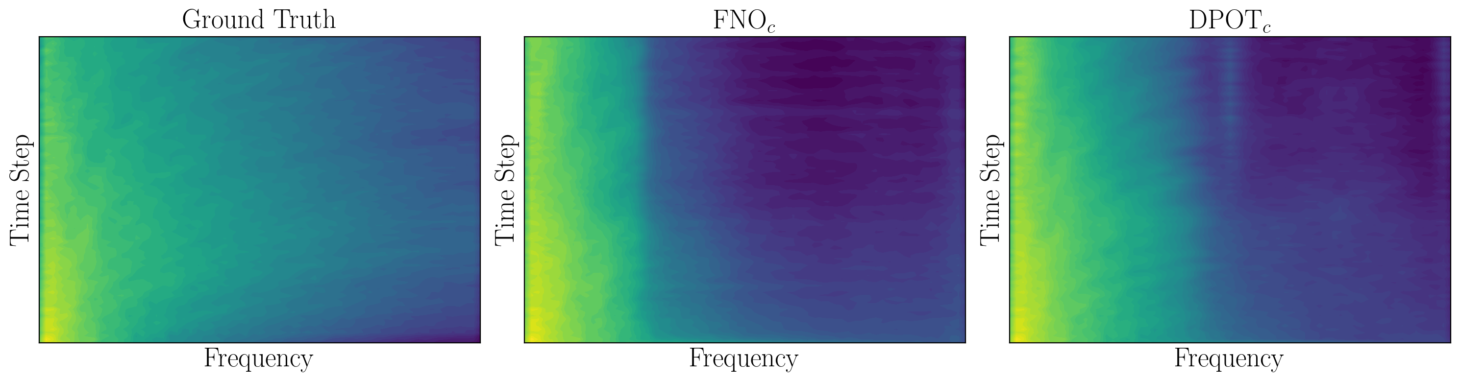}
    \caption{Log frequency density versus time for the FNO$_c$ and DPOT$_c$ models for the fluid density channel. While both models struggle with high frequency predictions, DPOT$_c$ exhibits higher frequency components than FNO$_c$.}
    \label{ch5.fig:2d_freq}
\end{figure}

A visualization of the TKE spectral density for the FNO and DPOT models for a sample from the Mach 0.1 dataset is displayed in Figure \ref{ch5.fig:tke}. Both model archetypes struggle to preserve high-frequency features, showing a lack of energy density at high frequencies compared to the ground truth data. However, the overall rate of energy decay from low frequencies to high frequencies is remarkably similar to the ground-truth, indicating that models are successfully representing the rate at which energy cascades. The lack of high-frequency representation is further supported by visualizing the density channel frequency spectra across time, presented in figure \ref{ch5.fig:2d_freq}. This visualization makes the models' inability to represent high-frequency data extremely apparent, with the FNO model exhibiting quantized cutoffs related to the number of Fourier modes. In contrast, the DPOT model is able to model higher frequency data than the number of Fourier modes in the model, though a significant cutoff is still observed at the highest of frequencies. This may be explained by the patch embedding approach of DPOT, which encode some high-frequency features prior to the frequency transform. Notably, however, a consequence of the patch decoding can also be observed, with a significant artifact appearing in the frequency spectra at high timesteps, correlating to the patch size of the DPOT operator. 

These results highlight a consistent and nontrivial limitation of present-day neural operators. While the conserved quantity correction methods presented in Section \ref{ch5.sec:cqc} help limit the effect of auto-regressive errors for long-term rollouts, all tested model archetypes still struggle to learn high-frequency data features, severely limiting their efficacy as high-frequency data becomes more prominent at later timesteps, as exhibiting in the ground truth spectral density in Figure \ref{ch5.fig:2d_freq}. Future work must consider additional methodologies to incorporate high-frequency components into the methodology if we are to further improve the applicability of neural operators for turbulent flow prediction.

\section{Conclusion}

In this work, we introduce conserved quantity correction, a methodology which may be generally applied to neural operator architectures. Our methods ensure that neural operators conserve fundamental quantities such as mass or momentum, reducing the effect of auto-regressive errors and stabilizing models during long-rollouts. Our results show that conserved quantity correction reduces the relative error of model predictions and significantly improves the correlation between model outputs and ground-truth data over long prediction durations. In future work, we aim to improve the ability of neural operators to represent high frequency data, and test the conserved quantity correction methodology on additional operator architectures and datasets.




\bibliography{references}

\end{document}